%% file: main_2027.tex
\documentclass[letterpaper]{article} 
\usepackage[preprint]{aaai2027}  
\usepackage[hyphens]{url}  
\usepackage{graphicx} 
\usepackage{natbib}  
\usepackage{caption} 
\usepackage{algorithm}
\usepackage{algorithmic}

\usepackage{newfloat}
\usepackage{listings}
\DeclareCaptionStyle{ruled}{labelfont=normalfont,labelsep=colon,strut=off} 
\floatstyle{ruled}
\newfloat{listing}{tb}{lst}{}
\floatname{listing}{Listing}

\usepackage{booktabs}
\usepackage{pifont}

\usepackage{amsmath}
\usepackage{amssymb}
\usepackage{bm}
\usepackage{booktabs}
\usepackage{multirow}
\usepackage{array}
\usepackage{textcomp}
\usepackage{colortbl}
\usepackage{xcolor}

\newcommand{\method}{\textsc{DLAM}}
\newcommand{\alam}{\textsc{ALAM}}

\newcommand{\KL}{D_{\mathrm{KL}}}
\newcommand{\diag}{\operatorname{diag}}
\newcommand{\N}{\mathcal{N}}
\newcommand{\cL}{\mathcal{L}}

\makeatletter
\def\input@path{{./}}
\makeatother

\title{\method{}: Distributional Latent Actions with Temporal Constraints}

\author{Zuojin Tang$^{1,2}$\quad Feifan Luo$^{1}$\quad  Haoyun Liu$^{3,5,2}$\quad  Botai Yuan$^{4,2}$\quad  Dekang Qi$^{2}$\quad \\Ronghan Chen$^{2}$\quad Yandan Yang$^{2}$\quad  Tong Lin$^{6,2}$\quad  Xinyuan Chang$^{2}$\quad \\Mu Xu$^{2}$\quad Bin Liu$^{7}$\quad  De Ma$^{1}$\quad Zhiheng Ma$^{5}$ }
\affiliations{$^1$Zhejiang University\quad $^2$Amap, Alibaba Group\quad $^3$Nanjing University\quad $^4$Shanghai Jiaotong University\quad\\ $^5$Shenzhen University of Advanced Technology\quad $^6$Xi'an Jiaotong University\quad \\$^7$Embodied Intelligence General Platform Laboratory, Chery Auto}

\begin{document}
\maketitle

\input{00_abs}
\input{01_intro}
\input{02_rw}
\input{03_met}

\input{04_exp}
\input{05_con}
\bibliography{aaai2027}


\end{document}

%% file: 00_abs.tex
\begin{abstract}
Vision-language-action (VLA) models remain constrained by scarce action-labeled robot data, whereas action-free videos offer abundant observations of physical change. Latent action models can extract such priors, but reconstruction-trained codes may predict future observations without the structure required for joint generation with robot actions. Existing structured methods add temporal constraints but retain deterministic transition points, so residual errors in locally inferred transitions may propagate and compound under recursive composition. We introduce \method{}, a distributional latent-action model that represents each transition as a diagonal Gaussian. Reconstruction conditioned on the reference frame grounds the mean in observed visual change, while normalized composition and reversal over equal-gap triplets constrain both the mean and dimension-wise variance. Variance composition uses a lightweight shared-correlation coefficient to account for dependence between adjacent transitions that share an intermediate frame, whereas reversal negates the mean and preserves the variance. For downstream policy learning, we freeze the encoder and train a flow-matching policy to jointly generate mean transition sequences and robot actions. On held-out transitions, \method{} learns more temporally consistent latent dynamics than existing latent-action baselines and achieves stronger direct and cumulative reconstruction on held-out videos. Under the same controlled $\pi_0$ transfer protocol, it also improves policy performance on MetaWorld MT50, LIBERO, and real-world manipulation tasks. Controlled ablations show that normalized mean constraints account for most of the reconstruction gain, while learned variance and correlation-aware composition provide complementary improvements in downstream control.

\end{abstract}

\begin{figure}[t]
    \centering
    \includegraphics[width=\linewidth]{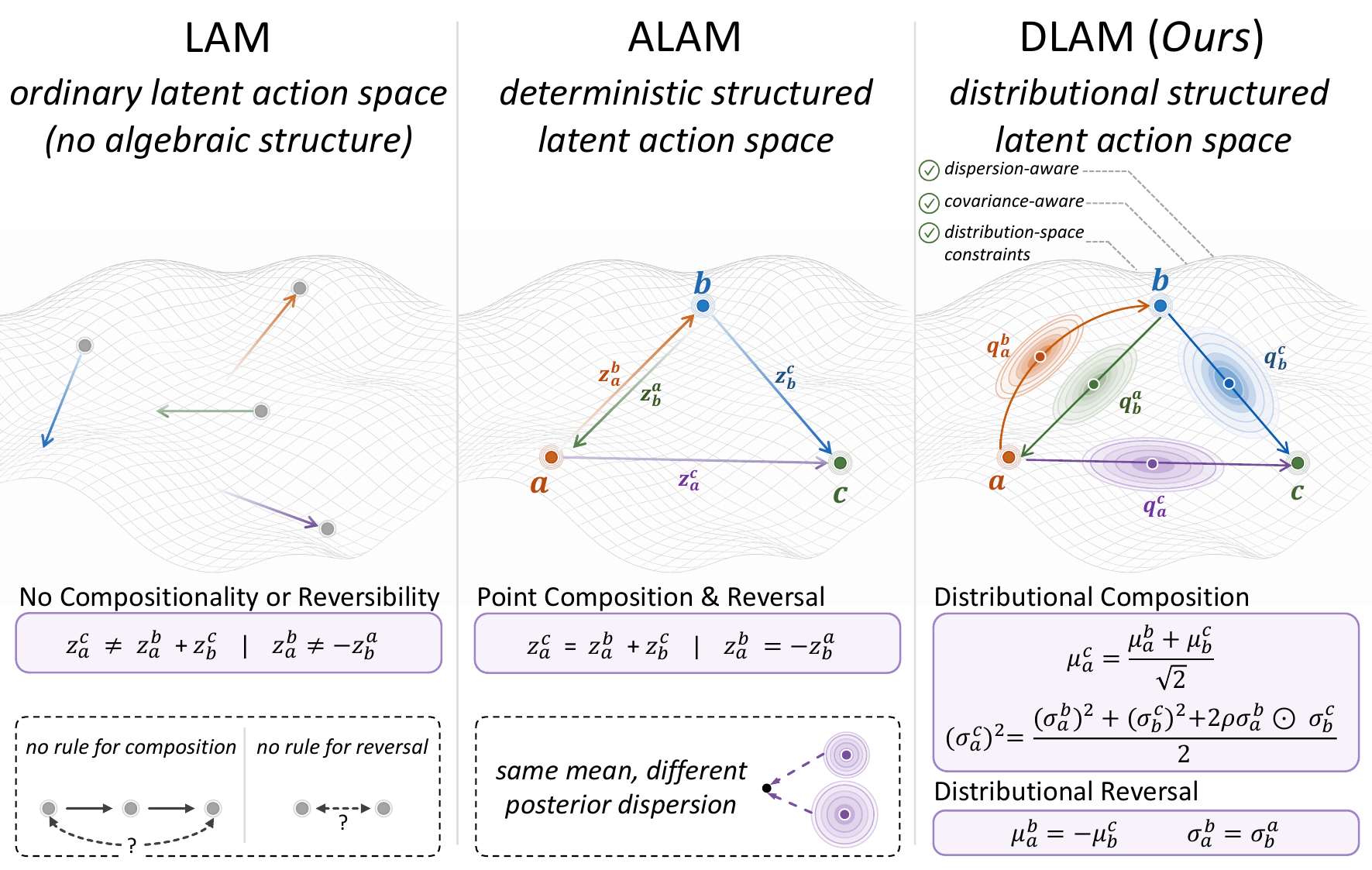}
    \caption{\textbf{From point-valued to distribution-valued transitions.}
    Reconstruction-only LAMs do not explicitly organize temporal relations.
    ALAM adds composition and reversal constraints to deterministic transition
    points. \method{} applies these relations to Gaussian posteriors, constraining
    both the transition center and its dimension-wise dispersion.}
    \label{fig:motivation}
\end{figure}

%% file: 01_intro.tex
\section{Introduction}

Vision-language-action (VLA) models provide a unified framework for visual perception, language understanding, and action generation in robot policies. Recent systems learn diverse manipulation skills from large robot datasets~\citep{zitkovich2023rt,kim2024openvla,black2024pi0,o2024open}, but their scalability remains limited by the cost of action-labeled demonstrations. Action-free videos, by contrast, provide abundant observations of physical change and offer a complementary source of transition priors for downstream control~\citep{chen2026abot,lingbot-va2026,shen2026world}.

Latent action models (LAMs) infer transitions without action labels, typically using reconstruction to ground the latent in observed change~\citep{bruce2024genie,ye2024latent,bu2025univla}. However, reconstruction alone does not guarantee a representation suitable for policy generation. The latent may encode camera motion, background dynamics, or appearance changes that improve frame prediction but are only weakly related to control~\citep{zhang2025whatlam}. Such a transition can be predictive without being well suited to joint generation with robot actions.

Structured LAMs impose composition, reversal, inverse, or cycle constraints~\citep{wei2026aclam,tang2026alam,li2026rotvla}, but still represent each transition as a deterministic point. Each relation thus acts on a single estimate, whose residual errors may compound under recursive long-horizon composition. We instead represent each transition as a diagonal Gaussian, allowing temporal constraints to supervise both its mean and dimension-wise variance. Reconstruction grounds the mean in observed change, while the variance provides an auxiliary signal to the shared encoder. 

We propose \method{}, a distributional latent-action model for action-free video. Given a frame pair, an encoder predicts the transition mean and dimension-wise variance, while a source-conditioned decoder reconstructs the target using only the mean. For an equally spaced triplet, normalized composition matches the directly encoded transition with the composition of two adjacent transitions in both mean and variance. As shown in Figure~\ref{fig:motivation}, a lightweight shared-correlation coefficient accounts for their dependence when composing variances. Reversal negates the mean while preserving the variance. These constraints are local and do not assume an exact global group structure. To transfer \method{} to downstream policies, we freeze the encoder and use its mean transition sequences as auxiliary generative targets alongside robot actions. A flow-matching policy jointly generates both trajectories, requiring neither the predicted variance, a latent-to-action decoder, nor a new VLA backbone.

We evaluate \method{} at the representation, reconstruction, and policy levels. Compared with existing latent-action baselines, \method{} achieves consistently lower composition and reversal residuals on held-out transitions, less sensitive to increasing temporal span under the common normalized probe. Normalized temporal probes further show that these residuals are less sensitive to increasing recursive span. \method{} also improves direct and cumulative reconstruction by $3.45\,\mathrm{dB}$ and $1.17\,\mathrm{dB}$, respectively, with consistent gains across complementary perceptual measures. Under the controlled $\pi_0$ transfer setting, the learned means achieve $87.6\%$ average success on MetaWorld MT50, $99.0\%$ on LIBERO, and $73.8\%$ across four real-world tasks. A controlled ablation further shows higher downstream success when introducing normalized mean constraints, learned variance, and correlation-aware composition. Our contributions are:
\begin{itemize}
    \item We extend structured latent actions from deterministic points to diagonal-Gaussian transitions, allowing temporal constraints to supervise both the mean and variance.
    \item We formulate normalized composition and reversal for Gaussian transitions, including correlation-aware composition of dimension-wise variances.
    \item We improve direct and cumulative reconstruction and transfer only the learned posterior means to joint flow-matching VLA learning. Controlled ablations further isolate the complementary contributions of normalized mean constraints, learned variance, and correlation-aware composition to downstream control.

\end{itemize}


\begin{figure*}[t]
    \centering
    \includegraphics[width=0.91\textwidth]{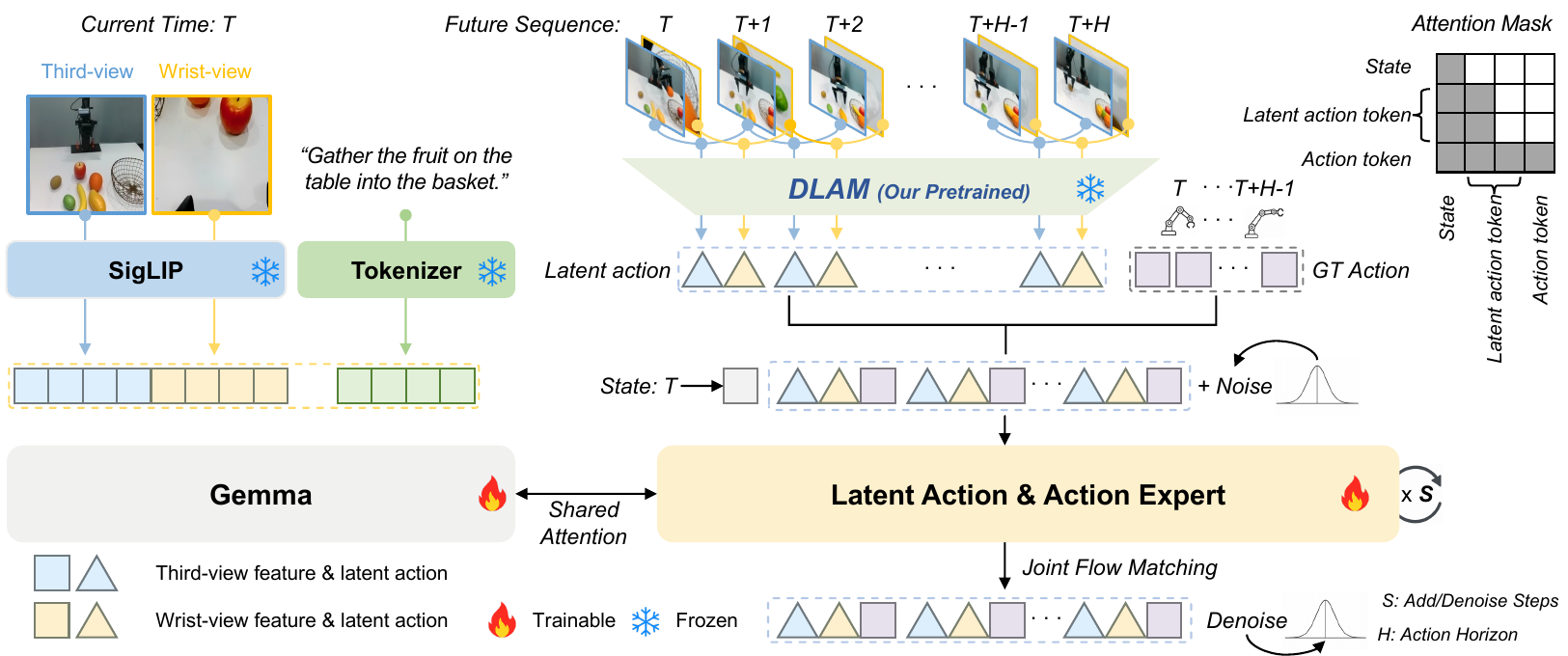}
    \caption{\textbf{Downstream transfer of \method{}.}
    The frozen encoder extracts distributional transitions from consecutive
    demonstration frames. Their posterior means are projected and interleaved
    with robot-action tokens. Conditioned on current visual observations and
    language, the policy jointly generates both streams with flow matching.
    Future frames are used only to construct training targets, and only robot
    actions are executed.}
    \label{fig:framework}
\end{figure*}

%% file: 02_rw.tex
\section{Related Work}

\subsection{Vision-Language-Action and World Action Models}

VLA models build on large-scale vision-language representations for robot
control. Representative systems include autoregressive models such as RT-2 and
OpenVLA, together with related large-scale VLA approaches
~\citep{zitkovich2023rt,kim2024openvla,tang2025vlascd,wang2026qwen,yuan2026qwen}. Diffusion Policy and $\pi_0$ instead generate continuous action trajectories through diffusion or flow matching
~\citep{chi2025diffusion,black2024pi0}. Cross-embodiment datasets further
broaden the tasks and robot morphologies available for policy learning
~\citep{o2024open}. Several methods also use future visual information to
guide action generation. Prediction with Action jointly denoises future
observations and actions, while video-prediction policies condition control on
predicted visual features~\citep{guo2024prediction,hu2024video}. World action
models more directly connect representations of visual dynamics with
executable actions ~\citep{tang2026one,chen2026abot,lingbot-va2026,yuan2026fast}. Our work follows this direction but focuses on the transition representation learned from action-free videos while keeping the downstream VLA backbone unchanged.

\subsection{Latent Action Models from Videos}

Latent action models infer action-like variables directly from observation
sequences. Genie learns discrete latent actions for interactive world
generation, while LAPA transfers video-derived latent actions to robot
policies~\citep{bruce2024genie,ye2024latent}. UniVLA, AdaWorld, and CLAM further study task-aware and continuous latent actions for cross-embodiment robot learning~\citep{bu2025univla,gao2025adaworld,liang2025clam}.
Reconstruction-based latents may, however, encode visual changes that help
prediction but are weakly related to control~\citep{zhang2025whatlam}.
AC-LAM, ALAM, and RotVLA address this issue by introducing explicit temporal
structure~\citep{wei2026aclam,tang2026alam,li2026rotvla}. Despite using
different operators, these methods represent each transition
deterministically. Stochastic video-prediction methods also use latent
variables, primarily to model multiple plausible futures
~\citep{babaeizadeh2018sv2p,denton2018svg}. In contrast, \method{} uses a
diagonal-Gaussian representation for the transition between two observed
frames and imposes temporal constraints on both its mean and variance. The variance provides an additional training signal rather than calibrated uncertainty, and downstream transfer uses only the learned mean.

%% file: 03_met.tex
\section{Method}
\label{sec:method}

\subsection{Overview}

\method{} extends deterministic latent-action learning with
diagonal-Gaussian transition tokens. Source-conditioned reconstruction grounds
each transition mean in the observed visual change, while equal-gap triplets
apply composition and reversal constraints to both the mean and variance,
providing an additional training signal beyond deterministic mean constraints.
For downstream control, the frozen encoder provides only mean targets, which
are jointly generated with robot actions (Figure~\ref{fig:framework}).

\subsection{Distributional Latent-Action Pretraining}
\label{sec:pretraining}

Let $O_t$ denote the video frame at time $t$. We sample equally spaced triplets
$(O_a,O_b,O_c)$ with $a<b<c$ and $b-a=c-b=k$, where $k$ is the frame gap. The
forward set
$\mathcal{T}_{\mathrm{fwd}}=\{(a,b),(b,c),(a,c)\}$ contains two adjacent
transitions and their direct counterpart, while the backward pair $(b,a)$
provides reversal supervision. No action, language, or proprioceptive labels
are required.

For an ordered pair $(O_i,O_j)$, let
$\bm Z_i^{\,j}=\{\bm Z_{i,\kappa}^{\,j}\}_{\kappa=1}^{K}$ denote the $K$
transition tokens inferred from source frame $O_i$ and target frame $O_j$. The
conditional posterior factorizes across token slots:
$q_i^{\,j}=q_\phi(\bm Z_i^{\,j}\mid O_i,O_j)
=\prod_{\kappa=1}^{K}q_{i,\kappa}^{\,j}$. Each factor is a diagonal Gaussian:
\begin{equation}
q_{i,\kappa}^{\,j}
=
\N\!\left(
\bm\mu_{i,\kappa}^{\,j},
\diag\!\big((\bm\sigma_{i,\kappa}^{\,j})^2\big)
\right).
\label{eq:transition_posterior}
\end{equation}
Here $\kappa$ indexes the token slot, $d$ is the token width, and
$\bm\mu_{i,\kappa}^{\,j}\in\mathbb{R}^{d}$ and
$\bm\sigma_{i,\kappa}^{\,j}\in\mathbb{R}_{+}^{d}$ are its mean and
element-wise standard deviation. Thus,
$\diag((\bm\sigma_{i,\kappa}^{\,j})^2)$ is the diagonal covariance, and $\N$
denotes a Gaussian distribution.

The relational encoder $E_\phi$, parameterized by $\phi$, processes the two
frames with $K$ learnable queries and predicts
$[\bm\mu_i^{\,j},\widetilde{\bm\ell}_i^{\,j}]=E_\phi(O_i,O_j)$, where both
outputs have shape $K\times d$. The raw log-variance
$\widetilde{\bm\ell}_i^{\,j}$ is converted to the standard deviation by
\begin{equation}
\bm\ell_i^{\,j}
=
\operatorname{clip}
(\widetilde{\bm\ell}_i^{\,j},\ell_{\min},\ell_{\max}),
\qquad
\bm\sigma_i^{\,j}
=
\exp\!\left(\tfrac{1}{2}\bm\ell_i^{\,j}\right).
\label{eq:transition_parameters}
\end{equation}
The clipping and exponential are applied element-wise, with
$\ell_{\min},\ell_{\max}$ as fixed bounds. Because both endpoints are observed
and the reconstruction path uses only the mean, the predicted standard
deviation is used to define the diagonal Gaussian and provide an additional
training signal through the prior and temporal constraints. It is not
interpreted as calibrated uncertainty about an unseen future.

For each $(i,j)\in\mathcal{T}_{\mathrm{fwd}}$, a source-conditioned decoder
$D_\omega$ reconstructs the target frame from the source frame and stacked
transition means:
$\widehat O^{\,i,j}=D_\omega(O_i,\bm\mu_i^{\,j})$. The reconstruction loss is
\begin{equation}
\cL_{\mathrm{rec}}
=
\frac{1}{P|\mathcal{T}_{\mathrm{fwd}}|}
\sum_{(i,j)\in\mathcal{T}_{\mathrm{fwd}}}
\left\|\widehat O^{\,i,j}-O_j\right\|_2^2,
\label{eq:reconstruction_loss}
\end{equation}
where $P$ is the number of scalar pixel and channel values in one frame, and
the norm is applied to its flattened representation. Conditioning on $O_i$
provides the scene content and encourages $\bm\mu_i^{\,j}$ to describe the
visual change toward $O_j$. No posterior sample is used in this reconstruction
path.

We regularize each forward posterior toward the factorized standard-Gaussian
prior $p=\prod_{\kappa=1}^{K}\N(\bm0,\bm I_d)$, where $\bm0$ and $\bm I_d$ are
the $d$-dimensional zero vector and identity matrix:
\begin{equation}
\cL_{\mathrm{prior}}
=
\frac{1}{|\mathcal{T}_{\mathrm{fwd}}|}
\sum_{(i,j)\in\mathcal{T}_{\mathrm{fwd}}}
\KL(q_i^{\,j}\|p).
\label{eq:prior_loss}
\end{equation}
Here $\KL$ denotes the Kullback--Leibler divergence. Because both distributions
factorize, the joint KL decomposes into the sum of $K$ token-wise Gaussian KL
terms. We apply the free-nats floor before the final loss reduction.

\begin{figure*}[t]
    \centering
    \includegraphics[width=0.8\linewidth]{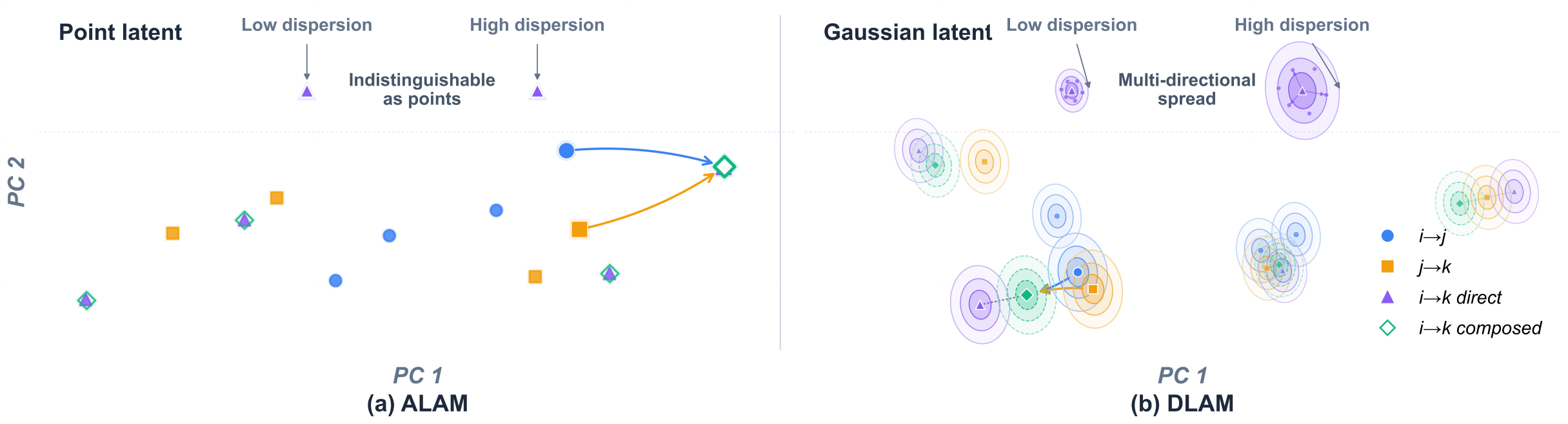}
    \caption{\textbf{Deterministic and diagonal-Gaussian transitions.}
    A PCA projection of held-out transitions encoded by trained ALAM and
    \method{} checkpoints. ALAM returns one point per transition, whereas
    \method{} predicts a mean together with an element-wise standard
    deviation. This visualization shows the difference in representation; it
    is not an uncertainty-calibration or cross-model geometric comparison.}
    \label{fig:latent_distribution}
\end{figure*}
\subsection{Temporal Constraints on Mean and Variance}
\label{sec:distributional_relations}

An equal-gap triplet provides adjacent posteriors $q_a^{\,b}$ and
$q_b^{\,c}$ together with the directly encoded posterior $q_a^{\,c}$.
Composition and reversal are applied independently to corresponding token
slots. For clarity, we fix a slot $\kappa$ and omit its index from the mean and
variance equations below.

Let $Z_{ab}\sim q_{a,\kappa}^{b}$ and
$Z_{bc}\sim q_{b,\kappa}^{c}$. We model the pair
$(Z_{ab},Z_{bc})$ as jointly Gaussian with these marginals
and define their normalized pairwise composition as
\begin{equation}
\bm Z_{a\rightsquigarrow c}
=
\frac{\bm Z_a^{\,b}+\bm Z_b^{\,c}}{\sqrt{2}}.
\label{eq:normalized_composition}
\end{equation}
Because the adjacent transitions share intermediate frame $O_b$, we allow a lightweight
coupling between them. A learned scalar $r\in\mathbb{R}$ defines
$\rho=\rho_{\max}\tanh(r)$, where the fixed bound
$0<\rho_{\max}<1$ ensures $|\rho|<1$. We share $\rho$ across examples, token
slots, and latent dimensions, and assume the diagonal cross-covariance
\begin{equation}
\operatorname{Cov}(\bm Z_a^{\,b},\bm Z_b^{\,c})
=
\diag\!\left(
\rho\,\bm\sigma_a^{\,b}\odot\bm\sigma_b^{\,c}
\right),
\label{eq:cross_covariance}
\end{equation}
where $\operatorname{Cov}$ denotes cross-covariance and $\odot$ denotes
element-wise multiplication.

The resulting composed mean and variance are
\begin{equation}
\overline{\bm\mu}_a^{\,c}
=
\frac{\bm\mu_a^{\,b}+\bm\mu_b^{\,c}}{\sqrt{2}},
\label{eq:composed_mean}
\end{equation}
\begin{equation}
(\overline{\bm\sigma}_a^{\,c})^2
=
\frac{(\bm\sigma_a^{\,b})^2+(\bm\sigma_b^{\,c})^2}{2}
+
\rho\,\bm\sigma_a^{\,b}\odot\bm\sigma_b^{\,c}.
\label{eq:composed_variance}
\end{equation}
The overbar distinguishes the composed mean and variance from the directly
encoded $\bm\mu_a^{\,c}$ and $(\bm\sigma_a^{\,c})^2$. It defines a pairwise normalized relation rather than an associative composition law. Restoring the token index, these quantities define
\begin{equation}
\overline q_{a,\kappa}^{\,c}
=
\N\!\left(
\overline{\bm\mu}_{a,\kappa}^{\,c},
\diag\!\big((\overline{\bm\sigma}_{a,\kappa}^{\,c})^2\big)
\right).
\label{eq:composed_posterior}
\end{equation}
Thus,
$\bm Z_{a\rightsquigarrow c,\kappa}\sim\overline q_{a,\kappa}^{\,c}$, and the
full composed posterior factorizes as
$\overline q_a^{\,c}
=\prod_{\kappa=1}^{K}\overline q_{a,\kappa}^{\,c}$.

Setting $\rho=0$ recovers independent variance propagation. In this case, the
$1/\sqrt{2}$ factor preserves unit variance when the two inputs are independent
standard Gaussians. We apply this operator once to the equal-gap relation
$k+k\!\rightarrow\!2k$; it defines a pairwise normalized relation rather than
an associative composition law.

Reversal changes the direction of a transition by negating its mean while
preserving its variance:
\begin{equation}
\mathcal{R}\!\left[
\N\!\left(\bm\mu,\diag\!\big((\bm\sigma)^2\big)\right)
\right]
=
\N\!\left(-\bm\mu,\diag\!\big((\bm\sigma)^2\big)\right).
\label{eq:reversal_operator}
\end{equation}
The operator is applied independently to all token slots. Applying it twice
recovers the original posterior. We use it as a forward--backward consistency
relation between $q_a^{\,b}$ and $q_b^{\,a}$, not as an exact inverse under
the normalized composition above.

To compare two factorized $K$-token posteriors $q$ and $q'$, let
$\bm\mu,\bm\mu',\bm\ell,\bm\ell'\in\mathbb{R}^{K\times d}$ denote their stacked
means and log-variances. We define the mean-and-log-variance discrepancy
\begin{equation}
\mathcal{D}(q,q')
=
\frac{
\|\bm\mu-\bm\mu'\|_F^2
+
\lambda_{\ell}\|\bm\ell-\bm\ell'\|_F^2
}{Kd},
\label{eq:posterior_discrepancy}
\end{equation}
where $\|\cdot\|_F$ is the Frobenius norm and $\lambda_{\ell}$ controls the
weight of log-variance matching. For the composed posterior,
$\overline{\bm\ell}_a^{\,c}
=\log((\overline{\bm\sigma}_a^{\,c})^2)$ is computed element-wise. The
temporal-constraint losses are
\begin{equation}
\cL_{\mathrm{comp}}
=
\mathcal{D}(q_a^{\,c},\overline q_a^{\,c}),
\label{eq:composition_loss}
\end{equation}

\begin{equation}
\cL_{\mathrm{rev}}
=
\mathcal{D}(q_a^{\,b},\mathcal{R}[q_b^{\,a}]).
\label{eq:reversal_loss}
\end{equation}

Composition matches the direct and composed posteriors, while reversal matches
the forward posterior with the transformed backward posterior.

Let
$\mathcal{S}=\{\mathrm{rec},\mathrm{prior},\mathrm{comp},\mathrm{rev}\}$.
The complete pretraining objective is
\begin{equation}
\cL_{\mathrm{DLAM}}
=
\sum_{s\in\mathcal{S}}\lambda_s\cL_s,
\label{eq:full_objective}
\end{equation}
where $\lambda_s$ weights the loss indexed by $s$.

The shared transition encoder jointly predicts and optimizes the mean and log-variance, giving the variance terms an additional training signal beyond reconstruction and the mean-based temporal constraints. Reconstruction and downstream policy transfer use only the resulting mean.

\subsection{Transfer to World Action Modeling}
\label{sec:transfer}

After pretraining, we discard the reconstruction decoder and freeze the
transition encoder. Let $I_t^m$ denote the frame at time $t$ from camera view
$m\in\{1,\ldots,M\}$, where $M$ is the number of views. Starting from the
current demonstration time $T$, the frozen encoder
$\mathcal{F}_\phi$ extracts transitions from $H$ consecutive frame pairs. For
$h=0,\ldots,H-1$,
\begin{equation}
(\bm\mu_h^m,\bm\ell_h^m)
=
\mathcal{F}_\phi(I_{T+h}^m,I_{T+h+1}^m).
\label{eq:downstream_transition}
\end{equation}
Here $\mathcal{F}_\phi$ includes the encoder and log-variance clipping in
Eq.~\eqref{eq:transition_parameters}; $h$ indexes a step within the $H$-step
target sequence, and both outputs have shape $K\times d$. Only
$\bm\mu_h^m$ is used as the downstream latent target; the log-variance is not
passed to the policy.

We use joint flow matching \citep{lipman2023flow,tang2026alam} to generate the
view-specific latent trajectories together with the executable robot-action
trajectory:
\begin{equation}
\cL_{\mathrm{transfer}}
=
\lambda_u\cL_{\mathrm{FM}}^{u}
+
\sum_{m=1}^{M}\lambda_m\cL_{\mathrm{FM}}^{m}.
\label{eq:transfer_objective}
\end{equation}
Here $\cL_{\mathrm{FM}}^{u}$ is the flow-matching loss for the executable action
stream, $\cL_{\mathrm{FM}}^{m}$ is the loss for the latent stream of view $m$,
and $\lambda_u,\lambda_m$ are their weights. Future demonstration frames and
the frozen encoder are used only to construct targets during policy training.
At inference time, the policy generates both streams, but only the robot-action
stream is executed.

%% file: 04_exp.tex
\section{Experiments}
\label{sec:experiments}

\subsection{Experimental Setup}
\label{sec:exp_setup}

\begin{figure}[t]
    \centering
    \includegraphics[width=\linewidth]{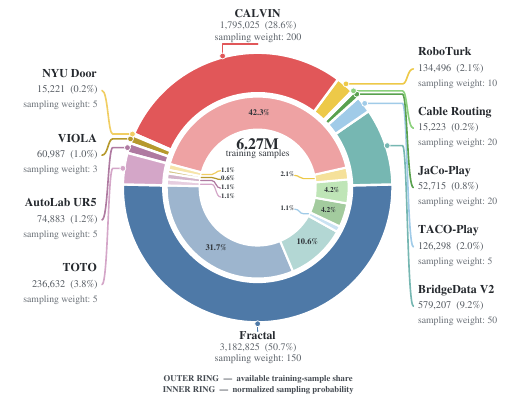}
    \caption{\textbf{Action-free pretraining mixture.}
    The outer ring shows the available training-sample share, whereas the
    inner ring shows the normalized sampling probability over the 11 video
    sources.}
    \label{fig:pre_training_datasets}
\end{figure}

\paragraph{Action-free pretraining.}
All latent-action models are pretrained on the same mixture of 11 action-free
robot-video datasets, drawn mainly from Open X-Embodiment
\citep{o2024open} and CALVIN \citep{mees2022calvin}.
Figure~\ref{fig:pre_training_datasets} shows the sampling weights. We use the
same visual tokenizer, Transformer capacity, source-conditioned decoder, data
order, and training budget for every controlled variant. Training runs for 57
epochs on 64 AMD MI308X GPUs with AdamW, a peak learning rate of
$10^{-4}$, weight decay of $10^{-4}$, and a per-device batch size of 64. We set
$\lambda_{\mathrm{rec}}=1$, $\lambda_{\mathrm{prior}}=0.005$,
$\lambda_{\mathrm{comp}}=\lambda_{\mathrm{rev}}=0.05$, and
$\lambda_{\ell}=0.1$.

\begin{table}[t]
\centering
\small
\setlength{\tabcolsep}{2.6pt}
\renewcommand{\arraystretch}{1.1}
\begin{tabular}{l c c | c c c c | c}
\toprule
\textbf{Method} & \textbf{Type} & \textbf{Size}
& \textbf{Easy} & \textbf{Med} & \textbf{Hard} & \textbf{V-Hard}
& \textbf{Avg.} \\
\midrule
RT-2              & AR & 7B   & 75.5 & 35.3 & 30.7 & 15.2 & 39.2 \\
RoboTron Mani     & AR & 4B   & 85.5 & 67.7 & 76.7 & 81.0 & 77.7 \\
\midrule
GR-1              & VA & 0.2B & 76.6 & 35.3 & 46.0 & 44.0 & 50.5 \\
PAD               & VA & --   & 81.8 & 65.1 & 56.7 & 87.2 & 72.7 \\
Evo-1             & VA & 0.8B & 89.2 & 76.8 & 77.2 & 79.2 & 80.6 \\
\midrule

$\pi_{0.5}$       & FM & 3B   & 68.2 & 37.3 & 41.7 & 28.0 & 43.8 \\
$\pi_0$           & FM & 3B   & 71.8 & 48.2 & 41.7 & 30.0 & 47.9 \\
SmolVLA           & FM & 2B   & 87.1 & 51.8 & 70.0 & 64.0 & 68.2 \\
$\pi_0$+\alam{}   & LA & 3B   & 89.3 & 83.6 & \textbf{85.0} & 82.0 & 85.0 \\
\midrule
\rowcolor{gray!12}
$\pi_0$+\method{} & LA & 3B
& \textbf{90.3} & \textbf{84.8}
& 84.0 & \textbf{91.3}
& \textbf{87.6} \\
\bottomrule
\end{tabular}
\caption{\textbf{Results on MetaWorld MT50.}
Avg.\ is the macro-average over the four difficulty tiers, computed before
rounding the displayed tier values. AR: autoregressive; FM: flow matching;
VA: video-augmented; LA: latent action.}
\label{tab:metaworld_main}
\end{table}

\begin{figure*}[t]
    \centering
    \includegraphics[width=0.97\textwidth]{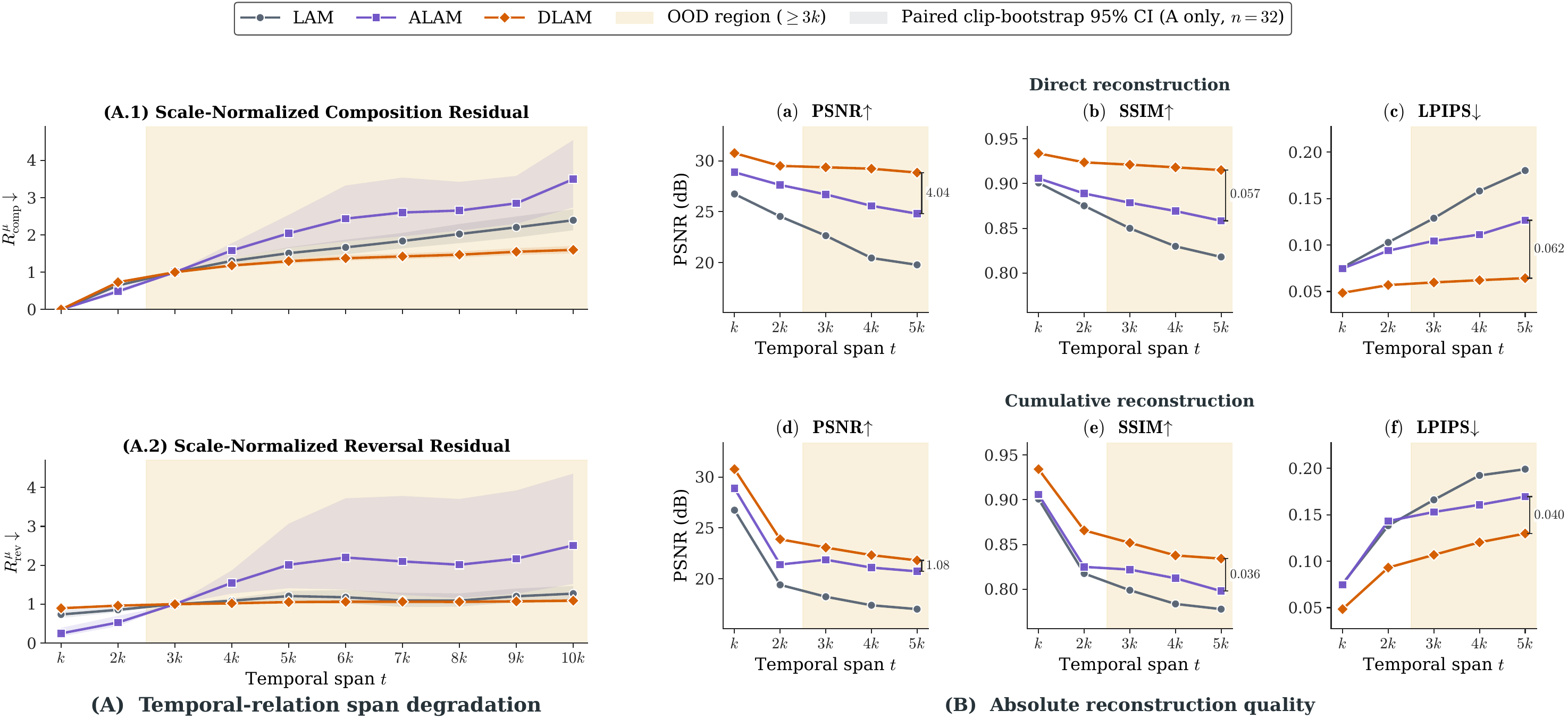}
    \caption{\textbf{Held-out transition diagnostics and reconstruction.}
    \textbf{A.} Scale-normalized mean composition and reversal residuals
    over increasing temporal spans. The shaded region marks spans beyond
    the directly supervised $k+k\rightarrow2k$ composition relation.
    \textbf{B.} Absolute reconstruction quality across endpoint spans for
    direct and cumulative decoding.}
    \label{fig:reconstruction_compare}
\end{figure*}

\paragraph{Policy transfer.}
We transfer each frozen transition encoder to the same $\pi_0$ policy
\citep{black2024pi0}, which uses PaliGemma-2B as the vision-language backbone
and Gemma-300M as the action expert \citep{team2024gemma}. For each
third-person and wrist-camera view, $H+1$ demonstration frames provide $H$
transition-mean targets. Only the policy backbone, action expert, and projection layers are updated. We train all variants with AdamW using a learning rate of $5\times10^{-5}$, weight decay of $10^{-4}$, and a per-device batch size of 32. More implementation details are provided in the Technical Supplement.

\paragraph{Benchmarks and baselines.}
On MetaWorld \citep{mclean2025metaworld}, we train one policy across 50
tasks, grouped into Easy, Medium, Hard, and Very Hard tiers. LIBERO
\citep{liu2023libero} evaluates manipulation generalization through its
Spatial, Object, Goal, and Long suites. We further evaluate on a Piper 6-DoF
arm using four real-world tasks: \emph{insert cylinder}, \emph{insert cube},
\emph{arrange flowers}, and \emph{hang cup}.

For transition representation and reconstruction, we compare with a
reconstruction-only LAM and deterministic ALAM \citep{tang2026alam}. The
MetaWorld table also includes RT-2 \citep{zitkovich2023rt}, RoboTron Mani
\citep{yan2024robotron}, GR-1 \citep{wu2023unleashing}, PAD
\citep{guo2024prediction}, Evo-1 \citep{lin2025evo}, $\pi_{0.5}$
\citep{intelligence2025pi_}, and SmolVLA \citep{shukor2025smolvla}. On
LIBERO, we additionally report OpenVLA \citep{kim2024openvla}, CoT-VLA
\citep{zhao2025cot}, DreamVLA \citep{zhang2025dreamvla}, OneWM-VLA
\citep{tang2026one}, GR00T-N1 \citep{bjorck2025gr00t}, LAPA
\citep{ye2024latent}, UniVLA \citep{bu2025univla}, and JALA
\citep{luo2026jointalignedlatentactionscalable}.

\subsection{Evaluating Transition Representations}
\label{sec:world_model_eval}

\paragraph{Evaluation protocol.}
We sample held-out windows from 11 datasets using $k=10$ frames per segment, evaluating reconstruction up to $5k$ and temporal relations up to $10k$. Direct reconstruction decodes the transition between the endpoints, whereas cumulative reconstruction composes adjacent $k$-frame transitions left to right before decoding.

For temporal diagnostics, $z$ denotes the posterior mean for \method{} and the deterministic embedding for point-valued baselines. We use the common length-aware probe
\begin{equation}
C_h(z_{1:h})=\frac{1}{\sqrt{h}}\sum_{i=1}^{h}z_i,
\label{eq:length_aware_probe}
\end{equation}
which matches \method{}'s training rule at $h=2$. For $h>2$, it measures per-length consistency rather than associativity. To reduce sensitivity to model-specific latent magnitudes, we define
\begin{equation}
R_{\mathrm{sym}}(x,y)=
\frac{\operatorname{mean}|x-y|}
{\frac{1}{2}\left(\operatorname{mean}|x|+\operatorname{mean}|y|\right)+\epsilon}.
\label{eq:symmetric_residual}
\end{equation}
We report
$R_{\mathrm{comp}}^{\mu}
=R_{\mathrm{sym}}(z_{\mathrm{direct}},C_h(z_{1:h}))$
and
$R_{\mathrm{rev}}^{\mu}
=R_{\mathrm{sym}}(z_{a\rightarrow b},-z_{b\rightarrow a})$.
Spans from $3k$ to $10k$ lie beyond the directly supervised
$k+k\rightarrow2k$ relation. Metrics are computed per window and then averaged; these mean-based probes do not test correlation-aware variance composition beyond the supervised span.

\begin{table}[t]
\centering
\small
\setlength{\tabcolsep}{2.6pt}
\renewcommand{\arraystretch}{1.1}
\begin{tabular}{l c c | c c c c | c}
\toprule
\textbf{Method} & \textbf{Type} & \textbf{Size}
& \textbf{Spatial} & \textbf{Object} & \textbf{Goal} & \textbf{Long}
& \textbf{Avg.} \\
\midrule
OpenVLA             & AR & 7B   & 84.7 & 88.4 & 79.2 & 53.7 & 76.5 \\
CoT-VLA             & AR & 7B   & 87.5 & 91.6 & 87.6 & 69.0 & 83.9 \\
\midrule
DreamVLA            & VA & 0.4B & 97.5 & 94.0 & 89.5 & 89.5 & 92.6 \\
OneWM-VLA           & VA & 3B   & 98.2 & 99.6 & 99.0 & 95.1 & 98.0 \\
\midrule
GR00T N1            & FM & 2B   & 94.4 & 97.6 & 93.0 & 90.6 & 93.9 \\
$\pi_0$             & FM & 3B   & 96.8 & 98.8 & 95.8 & 85.2 & 94.1 \\
$\pi_{0.5}$         & FM & 3B   & 98.8 & 98.2 & 98.0 & 92.4 & 96.9 \\
\midrule
LAPA                & LA & 7B   & 87.4 & 91.2 & 90.0 & 65.4 & 83.5 \\
UniVLA              & LA & 9B   & 96.5 & 96.8 & 95.6 & 92.0 & 95.2 \\
JALA                & LA & 3B   & 96.0 & 98.2 & 97.4 & 96.0 & 96.9 \\
$\pi_0$+\alam{}     & LA & 3B   & 99.2 & 99.6 & 99.0 & 94.4 & 98.1 \\
\midrule
\rowcolor{gray!12}
$\pi_0$+\method{}   & LA & 3B
& \textbf{99.6} & \textbf{99.8}
& \textbf{99.6} & \textbf{97.1}
& \textbf{99.0} \\
\bottomrule
\end{tabular}
\caption{\textbf{Success rates on LIBERO.}
Avg.\ is the mean over the four suites, computed before rounding the displayed
suite values. AR: autoregressive; FM: flow matching; VA: video-augmented;
LA: latent action.}
\label{tab:libero_main}
\end{table}

\paragraph{Latent transition visualization.}
Figure~\ref{fig:latent_distribution} illustrates the difference between the learned representations. For a fair comparison, we use checkpoints trained for 57 epochs under the same protocol for both methods. ALAM maps each transition to a single point, whereas \method{} predicts a mean and a diagonal variance vector. This visualization shows the difference in representation and is not an uncertainty-calibration analysis.

\paragraph{Temporal-span diagnostics.}
Figure~\ref{fig:reconstruction_compare}(A) reports composition and reversal
residuals of the learned latent transitions over progressively longer
temporal spans, using the length-aware probe in
Eq.~\eqref{eq:length_aware_probe}. Across the unseen $3k$--$10k$ range,
residuals for \method{} stay low and degrade only mildly with span, while
those for ALAM increase markedly. This pattern is consistent with improved
approximate temporal consistency under the reported diagnostic, though it
reflects behavior on the evaluated spans rather than a general guarantee at
arbitrary horizons.

\begin{figure*}[t]
    \centering
    \includegraphics[width=0.9\textwidth]{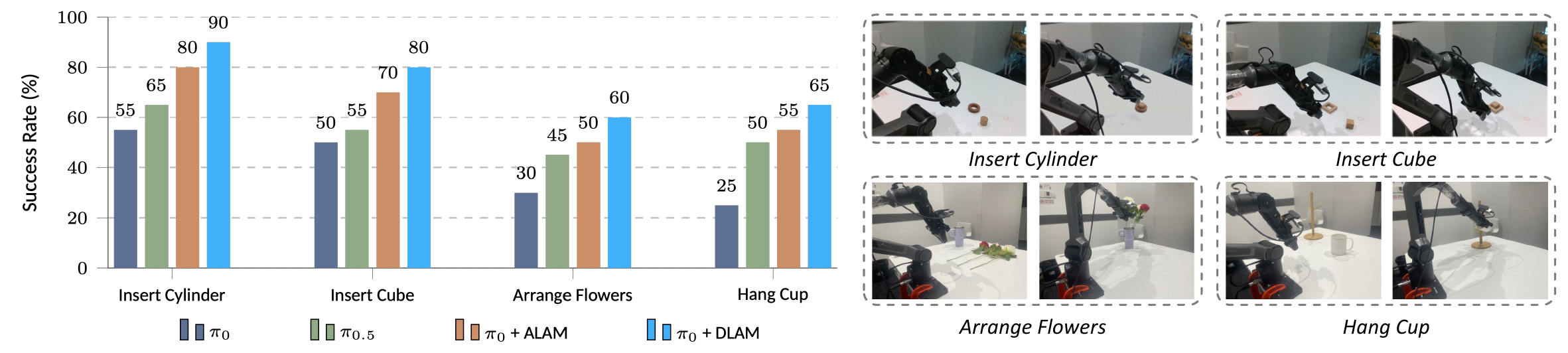}
    \caption{\textbf{Real-world manipulation.}
    Task-level success rates on a Piper 6-DoF arm for \emph{insert cylinder},
    \emph{insert cube}, \emph{arrange flowers}, and \emph{hang cup}. We compare
    action-only and latent-action-augmented policies under the same
    training and evaluation.}
    \label{fig:real_world_results}
\end{figure*}

\begin{table*}[t]
\centering
\small
\setlength{\tabcolsep}{6pt}
\renewcommand{\arraystretch}{1.12}

\begin{tabular}{@{}lcccccc@{}}
\toprule
&
\multicolumn{3}{c}{\textbf{Reconstruction Quality}}
&
\multicolumn{2}{c}{\textbf{Scale-Normalized Relations}}
&
\multicolumn{1}{c}{\textbf{Downstream Control}}
\\
\cmidrule(lr){2-4}
\cmidrule(lr){5-6}
\cmidrule(lr){7-7}

\textbf{Variant}
& \textbf{PSNR} $\uparrow$
& \textbf{SSIM} $\uparrow$
& \textbf{LPIPS} $\downarrow$
& $R_{\mathrm{comp}}^{\mu}$ $\downarrow$
& $R_{\mathrm{rev}}^{\mu}$ $\downarrow$
& \shortstack{\textbf{Avg.Success (\%)} $\uparrow$}
\\
\midrule

No temporal relations
& 21.236
& 0.7894
& 0.1755
& 1.2071
& 1.0314
& 76.6
\\

\shortstack[l]{Matched mean-only ($\sigma\!=\!1,\rho\!=\!0$)}
& 22.109
& 0.8057
& 0.1644
& 1.1808
& 1.0371
& 82.1
\\

Learned variance ($\rho\!=\!0$)
& 22.084
& 0.8067
& 0.1637
& 1.1777
& 1.0341
& 85.3
\\

\textbf{Full \method{}}
& \textbf{22.400}
& \textbf{0.8086}
& \textbf{0.1623}
& \textbf{1.1662}
& \textbf{1.0010}
& \textbf{87.6}
\\

\bottomrule
\end{tabular}

\caption{\textbf{Controlled ablation of \method{}.}
Held-out reconstruction and relation results are averaged over
$3k$--$5k$ temporal spans. Reconstruction uses cumulative decoding.
Downstream performance reports
the macro-average success rate on MetaWorld.}

\label{tab:dlam_ablation}
\end{table*}

\paragraph{Direct and cumulative reconstruction.}
We next evaluate whether the learned transition mean remains informative for
reconstruction across larger endpoint spans. Across the $3k$--$5k$ spans,
\method{} achieves average PSNR values of $29.14\,\mathrm{dB}$ for direct
reconstruction and $22.40\,\mathrm{dB}$ for cumulative reconstruction,
improving over ALAM by $3.45\,\mathrm{dB}$ and $1.17\,\mathrm{dB}$,
respectively (Figure~\ref{fig:reconstruction_compare}(B)). The same trend
holds for the perceptual metrics: \method{} reduces LPIPS by $45.6\%$ for
direct reconstruction and by $26.1\%$ after cumulative composition, with
consistent improvements in SSIM. These results show that the learned mean retains information useful for endpoint reconstruction across the evaluated spans.

\subsection{Downstream Policy Learning}
\label{sec:downstream_transfer}

Table~\ref{tab:metaworld_main} reports the results on MetaWorld MT50. Under the
controlled $\pi_0$ transfer setting, \method{} achieves 87.6\% average success,
improving the deterministic transition baseline by 2.6 percentage points and
the action-only policy by 39.7 points. The largest tier-level gain over ALAM
occurs on Very-Hard, where success increases from 82.0\% to 91.3\%.
As shown in Table~\ref{tab:libero_main}, \method{} reaches 99.0\% average
success on LIBERO, with improvements across all four suites. The largest gain
is observed on LIBERO-Long, where it improves the deterministic baseline by
2.7 percentage points. Figure~\ref{fig:real_world_results} further evaluates
the same transfer strategy on four real-world manipulation tasks.
$\pi_0+\method{}$ achieves the highest success rate on every task and reaches
an average of 73.8\%, compared with 63.8\% for $\pi_0+\mathrm{ALAM}$, 53.8\%
for $\pi_{0.5}$, and 40.0\% for $\pi_0$. In particular, it improves over the
deterministic transition baseline by 10 percentage points on each task.

Together, these results show that pretraining with constraints on transition
means and variances produces mean representations that transfer effectively
under the reported simulation and real-world evaluation protocols. They
establish the utility of the complete training formulation without attributing
the gains to calibrated uncertainty, learned variance alone, or the
temporal-span diagnostic.

\subsection{Ablation Studies}
\label{sec:ablations}

We compare four configurations using the same data, architecture, reconstruction objective, and training budget. \emph{No temporal relations} retains the Gaussian encoder and prior but removes composition and reversal. \emph{Matched mean-only} fixes $\sigma=1$ and $\rho=0$ while retaining the normalized mean composition, reversal constraint, and mean-prior penalty of the full model. \emph{Learned variance ($\rho=0$)} restores the predicted variance but propagates adjacent transitions independently, whereas the full \method{} additionally learns the shared coupling coefficient $\rho$. As shown in Table~\ref{tab:dlam_ablation}, the normalized mean
constraints recover most of the reconstruction gain, increasing
cumulative PSNR from $21.236$ to $22.109$\,dB and downstream success
from $76.6\%$ to $82.1\%$. Learning the variance with $\rho=0$
maintains comparable reconstruction quality, slightly reduces both
relation residuals, and further improves success to $85.3\%$.
Introducing the shared coupling coefficient yields the best results
across all metrics: full \method{} reaches $22.400$\,dB PSNR,
composition and reversal residuals of $1.1662$ and $1.0010$, and
$87.6\%$ downstream success, an $11.0$-point improvement over the
variant without temporal relations. These results isolate the
complementary contributions of normalized mean constraints, learned
variance, and correlation-aware composition.

%% file: 05_con.tex
\section{Conclusion}
\label{sec:conclusion_and_discussion}


We presented DLAM, a distributional latent-action model that represents visual transitions as diagonal Gaussians, applying normalized composition and reversal constraints to both the means and the dimension-wise variances. Under a common scale-normalized probe, the learned means show improved temporal consistency and retain more information for both direct and cumulative reconstruction. Controlled ablations reveal that normalized mean constraints drive most of this reconstruction improvement, while learned variance and correlation-aware composition contribute complementary gains in downstream control. The full formulation performs best across reconstruction quality, relational diagnostics, and policy transfer. In downstream learning, the frozen encoder contributes only posterior-mean transition sequences as auxiliary flow-matching targets, allowing the learned representation to enhance control without altering the policy backbone or execution interface.

\paragraph{Limitations.}
DLAM imposes local constraints on equal-gap triplets, leaving long-horizon generalization open. The variance objective may admit near-constant solutions. Since downstream transfer uses only posterior means, variance serves as an auxiliary training signal rather than calibrated uncertainty. The shared correlation may also miss context- or dimension-dependent dependencies.

